\documentclass{article} 
\usepackage{iclr2015,times}
\usepackage{hyperref}
\usepackage{url}

\usepackage{graphicx}
\usepackage{caption}
\usepackage{subcaption}

\title{Learning Compact Convolutional Neural Networks with Nested Dropout}

\author{
Chelsea Finn, Lisa Anne Hendricks \& Trevor Darrell\\
Department of Computer Science\\
UC Berkeley\\
Berkeley, CA 94704, USA \\
\texttt{\{cbfinn, lisa\_anne, trevor\}@eecs.berkeley.edu} \\
}

\iclrfinalcopy 

\begin{document}

\maketitle

\begin{abstract}
Recently, nested dropout was proposed as a method for ordering representation units in autoencoders by their information content, without diminishing reconstruction cost \citep{nesteddropout}. 
However, it has only been applied to training fully-connected autoencoders in an unsupervised setting. 
We explore the impact of nested dropout on the convolutional layers in a CNN trained by backpropagation, investigating whether nested dropout can provide a simple and systematic way to determine the optimal representation size with respect to the desired accuracy and desired task and data complexity. 
\end{abstract}

\section{Motivation}
Supervised convolutional neural networks (CNNs) learn representations that are effective on a wide range of tasks.
A drawback of current such approaches, however, is that the selection of such architectures is largely optimized by hand, with researchers explicitly searching architecture hyperparameters via cross-validation.  
Model selection for network architectures has been explored in the context of learning network connectivity dating back to Optimal Brain Damage from ~\citet{lecun1989optimal} and has been continued to be explored in the context of learning optimal sparse models.  
To our knowledge there is no deep visual network capable of increasing its representation capacity based on the complexity of available data or tasks. 
The recently proposed nested dropout method implicitly accomplishes this, by learning deep representation units in an incremental fashion.  
We investigate whether such an approach is applicable to visual CNN models, and propose a visual CNN model which can learn to scale its capacity according to the complexity of the data presented to the network during training.

In standard dropout, units in the layer are independently dropped out with probability $p$, namely the output of that unit is set to zero. 
This is traditionally applied to convolutional and fully-connected layers during training time, and has been shown to act as a regularizer, discouraging over-fitting to the training data \citep{dropout}, though it has also been applied to entire channels of a convolutional layer output by \citet{spatialdropout}.
Empirically, when training large networks, such as those trained on ImageNet, drop out is necessary to avoid over fitting~\citep{krizhevsky2012imagenet}.
Nested dropout, on the other hand, randomly draws unit indices from a geometric distribution and drops out all of the units that follow the number drawn, e.g. if a number $k$ is drawn, then the units $0$ through $k$ are kept and the remaining units are dropped. 
When applied to a single layer semi-linear autoencoder, this technique has been proven to enforce an ordering of the units by their information capacity, while not decreasing the flexibility of the representation nor the quality of the resulting solution~\citep{nesteddropout}.


The primary contributions of this paper are to (1) demonstrate that nested dropout can successfully be applied to convolutional layers trained by back-propagation, (2) propose nested dropout as an advantageous method to learn CNNs that adapt to task and data complexity in a deep learning setting, and (3) provide our implementation in Caffe \citep{jia2014caffe}, a widely used deep learning framework, upon publication.

\section{Nested Dropout on CNNs}
The nested dropout algorithm for a convolutional layer with $n$ channels is as follows: for each sample in a mini-batch, we draw a number $k$ from a geometric distribution and drop out the latter $n-k$ channels of the output of the layer.

Nested dropout can also be applied to multiple layers in a network by applying nested dropout to each layer iteratively.
First, the number of filters $n_i$ for layer $i$ is determined through nested dropout.  
After fixing the number of filters $n_i$ in layer $i$, nested dropout can then be used to determine the number of filters, $n_{i+1}$, for layer $i+1$.

We trained our CNNs using Stochastic Gradient Descent (SGD) and mini-batches of 100 samples. 
Because dropped out units are determined by drawing from a geometric distribution, units with a low index are rarely dropped out and thus converge quickly, whereas latter units are frequently dropout out and thus learned very slowly.
Filters are incrementally fixed once they have converged, and only the remaining filters are considered when drawing numbers
from the geometric distribution.
Though incrementing the sweeping index could be done upon filter convergence, we achieved satisfactory results by simply incrementing the unit sweeping index after a set number of iterations.

To implement this in the Caffe framework, we added a nested dropout layer that can follow any layer (e.g. convolutional, fully-connected) in the same way as the standard dropout layer. We also added customizations to the solver to support unit sweeping during training.

\section{Experiments}

\begin{figure}[h]
\begin{center}
\includegraphics[width=0.75\linewidth]{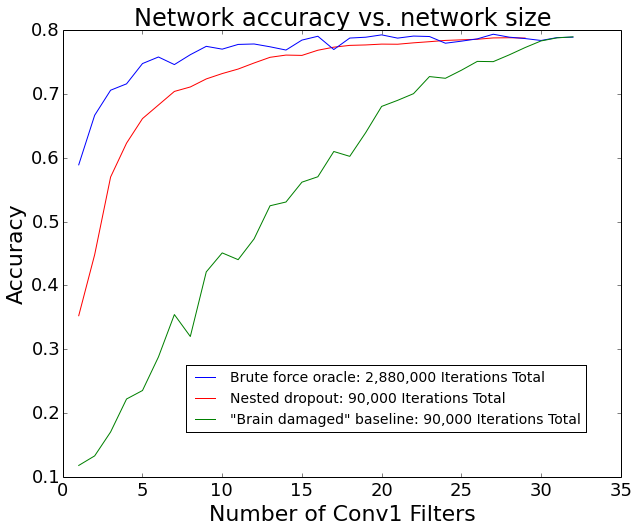}
\end{center}
\caption{Accuracy as a function of the number of filters in conv1, with and without nested dropout.  The oracle consists of 32 separate networks, whereas the nested dropout curve only requires training a single network.   The ``brain damaged'' baseline is trained without nested dropout and only the $k$ best filters are used during test.  The nested dropout network achieves maximum accuracy with a compact representation of 23 filters while requiring considerably fewer iterations than a brute force approach.}

\label{fig:plot}
\end{figure}
We apply nested dropout to the first convolutional layer of a CNN trained to classify images in the CIFAR-10 dataset, using the default Caffe architecture and training with a fixed learning rate. 
In Figure~\ref{fig:plot}, we show the test accuracy of a single network trained with nested dropout as a function of the number of conv1 filters.  
We report the accuracy with $k$ filters by using the partially-trained network after the $k$th filter has converged, and testing it with the last $n-k$ filters dropped out.
We compare to two naive approaches to select the number of filters in conv1.  The first approach simply trains 32 separate networks which differ in the number of conv1 filters.  The second approach trains a single network with 32 filters, but at test time, a varying number of conv1 filters are used with the remaining dropped out.  Note that because there is no inherent ordering to the learned representation without nested dropout, removing a filter severely damages the network resulting in low accuracies.  

Our experiments in Figure~\ref{fig:plot} demonstrate that nested dropout efficiently determines the relationship between model capacity and test accuracy.   The nested dropout implementation requires 90,000 iterations of training, whereas the brute force approach requires significantly more, up to millions iterations. For example, training $32$ separate networks to completion requires 2,880,000 iterations.

\begin{figure}[h]
\begin{center}
\begin{subfigure}{.5\textwidth}
  \centering
  \includegraphics[width=.8\linewidth]{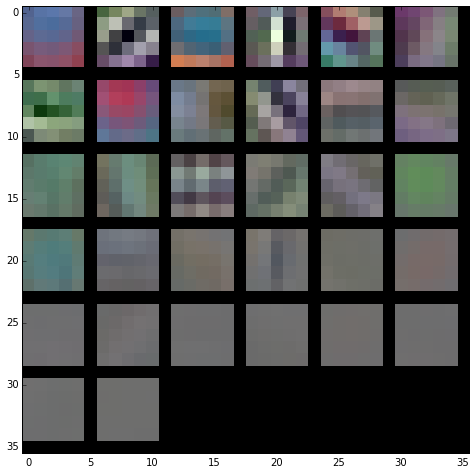}
  \caption{}
  \label{fig:sub1}
\end{subfigure}%
\begin{subfigure}{.5\textwidth}
  \centering
  \includegraphics[width=.8\linewidth]{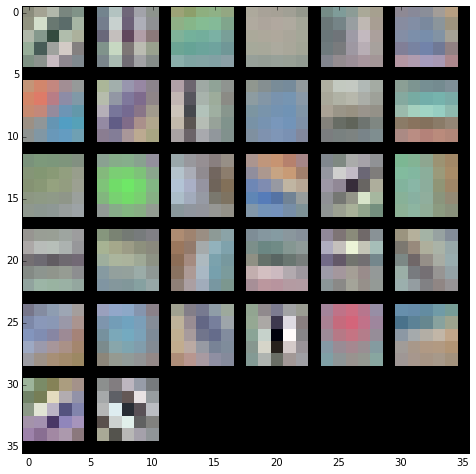}
  \caption{}
  \label{fig:sub2}
\end{subfigure}
\end{center}
\caption{conv1 filters of networks trained with nested dropout (a) and without (b). Note that the latter filters of (a) carry very little information, yet both networks achieve similar performance, giving an indication for how many units are necessary for this model.}
\label{fig:conv}
\end{figure}

In Figure~\ref{fig:conv}, we visualize the filters learned with and without nested dropout. Though the latter filters carry very little information, the test accuracy of the two sets of filters are comparable with 0.787 test accuracy for the network trained with nested dropout and 0.786 test accuracy for the baseline network.

Applying nested dropout to the second convolutional layer yielded similar results to our experiments with conv1.
After training a network with nested dropout applied to conv1 filters, we determined only 23 conv1 filters are necessary to achieve the maximum classification accuracy for this network.
We next train a network in which nested dropout follows conv2 and with a unit sweeping index of 5,000.
After learning 25 conv2 filters, the training accuracy converges to 78\%.
Thus, by using nested dropout we can reduce the total number of parameters in the first two layers by 25\% from 64 filters to 48, while maintaining similar accuracy.

\section{Discussion}

In summary, we have provided a simple method for determining a more compact representation of the convolutional layers. In our experiments, we learned a representation that achieved the same classification accuracy using 23 conv1 filters and 25 conv2 filters rather than the baseline 32 each, within the same optimization framework. A main advantage of our method is that it enables the network to gradually increase network capacity during training.    Additionally, we hope that, in the future, ordering parameters may provide insights into optimization of deep convolutional neural networks and how the network architecture impacts performance.

\subsubsection*{Acknowledgments}
This work was supported in part by DARPA's MSEE and SMISC programs, NSF awards IIS-1427425, IIS-1212798, and IIS-1116411, Toyota, and the Berkeley Vision and Learning Center. Chelsea Finn was supported by a Berkeley EECS Fellowship and Lisa Anne Hendricks by
an NDSEG Fellowship.

\bibliography{iclr2015}

\begin{thebibliography}{6}
\providecommand{\natexlab}[1]{#1}
\providecommand{\url}[1]{\texttt{#1}}
\expandafter\ifx\csname urlstyle\endcsname\relax
  \providecommand{\doi}[1]{doi: #1}\else
  \providecommand{\doi}{doi: \begingroup \urlstyle{rm}\Url}\fi

\bibitem[Hinton et~al.(2012)Hinton, Srivastava, Krizhevsky, Sutskever, and
  Salakhutdinov]{dropout}
Hinton, Geoffrey~E., Srivastava, Nitish, Krizhevsky, Alex, Sutskever, Ilya, and
  Salakhutdinov, Ruslan.
\newblock Improving neural networks by preventing co-adaptation of feature
  detectors.
\newblock \emph{CoRR}, abs/1207.0580, 2012.
\newblock URL \url{http://arxiv.org/abs/1207.0580}.

\bibitem[Jia et~al.(2014)Jia, Shelhamer, Donahue, Karayev, Long, Girshick,
  Guadarrama, and Darrell]{jia2014caffe}
Jia, Yangqing, Shelhamer, Evan, Donahue, Jeff, Karayev, Sergey, Long, Jonathan,
  Girshick, Ross, Guadarrama, Sergio, and Darrell, Trevor.
\newblock Caffe: Convolutional architecture for fast feature embedding.
\newblock In \emph{Proceedings of the ACM International Conference on
  Multimedia}, pp.\  675--678. ACM, 2014.

\bibitem[Krizhevsky et~al.(2012)Krizhevsky, Sutskever, and
  Hinton]{krizhevsky2012imagenet}
Krizhevsky, Alex, Sutskever, Ilya, and Hinton, Geoffrey~E.
\newblock Imagenet classification with deep convolutional neural networks.
\newblock In \emph{Advances in neural information processing systems}, pp.\
  1097--1105, 2012.

\bibitem[LeCun et~al.(1989)LeCun, Denker, Solla, Howard, and
  Jackel]{lecun1989optimal}
LeCun, Yann, Denker, John~S, Solla, Sara~A, Howard, Richard~E, and Jackel,
  Lawrence~D.
\newblock Optimal brain damage.
\newblock In \emph{NIPs}, volume~2, pp.\  598--605, 1989.

\bibitem[Rippel et~al.(2014)Rippel, Gelbart, and Adams]{nesteddropout}
Rippel, Oren, Gelbart, Michael~A, and Adams, Ryan~P.
\newblock Learning ordered representations with nested dropout.
\newblock \emph{arXiv preprint arXiv:1402.0915}, 2014.

\bibitem[Tompson et~al.(2014)Tompson, Goroshin, Jain, LeCun, and
  Bregler]{spatialdropout}
Tompson, Jonathan, Goroshin, Ross, Jain, Arjun, LeCun, Yann, and Bregler,
  Christoph.
\newblock Efficient object localization using convolutional networks.
\newblock \emph{CoRR}, abs/1411.4280, 2014.
\newblock URL \url{http://arxiv.org/abs/1411.4280}.

\end{thebibliography}
\bibliographystyle{iclr2015}

\end{document}